# Predicting Confinement Effect of Carbon Fiber Reinforced Polymers on Strength of Concrete using Metaheuristics-based Artificial Neural Networks

**Sarmed Wahab [1], Mohamed Suleiman [2], Faisal Shabbir [1], Nasim Shakouri Mahmoudabadi[3], Sarmad Waqas[4], Nouman Herl[5], Afaq Ahmad[3]**

[1]Department of Civil Engineering, University of Engineering and Technology, Taxila, 47080, Pakistan

[2]Civil Engineering Department, Tripoli University, Tripoli, Libya

[3]Department of Civil Engineering, The University of Memphis, TN 38152, USA;

[4]Project Manger. Osmani & Compnay Pvt. Ltd. Pakistan  sarmad.waqas@gmail.com

[5]Federal Board of Intermediate and Secondary Education, Pakistan noumanherl@hotmail.com

* **Correspondence:**
Afaq Ahmad
aahmad4@memphis.edu

**Keywords:** carbon fiber reinforced polymer, concrete, confinement effect, strength, particle swarm optimization, grey wolf optimizer, bat algorithm

## Abstract

This article deals with the study of predicting the confinement effect of carbon fiber reinforced polymers (CFRPs) on concrete cylinder strength using metaheuristics-based artificial neural networks. A detailed database of 708 CFRP confined concrete cylinders is developed from previously published research with information on 8 parameters including geometrical parameters like the diameter ($d$) and height ($h$) of a cylinder, unconfined compressive strength of concrete ($f'_{co}$), thickness ($nt$), the elastic modulus of CFRP ($E_f$), unconfined concrete strain ($\varepsilon_{co}$), confined concrete strain ($\varepsilon_{cc}$) and the ultimate compressive strength of confined concrete ($f'_{cc}$). Three metaheuristic models are implemented including particle swarm optimization (PSO), grey wolf optimizer (GWO), and bat algorithm (BA). These algorithms are trained on the data using an objective function of mean square error and their predicted results are validated against the experimental studies and finite element analysis. The study shows that the hybrid model of PSO predicted the strength of CFRP-confined concrete cylinders with maximum accuracy of 99.13% and GWO predicted the results with an accuracy of 98.17%. The high accuracy of axial compressive strength predictions demonstrated that these prediction models are a reliable solution to the empirical methods. The prediction models are especially suitable for avoiding full-scale time-consuming experimental tests that make the process quick and economical.

## 1 Introduction

Fiber-reinforced polymer is a composite material comprising fibers of either glass, aramid, or carbon and a polymer matrix. These fibers improve the properties of the polymer matrix mechanically including its stiffness and strength. The popularity of these composites has increased significantly in civil engineering due to their ability to strengthen concrete structural members. FRPs can be used either as a bar or plates embedded in concrete as an internal reinforcement and can be used as an external reinforcement by wrapping FRP sheets to existing structural members. The FRP bars have significantly higher strength than the steel reinforcement bars. They are highly durable and resistant to chemicals, corrosion (Cousin et al. 2019, Ananthkumar et al. 2020, Zhang et al. 2020), and radiation, their higher strength-to-weight ratio (Zhou et al. 2019) makes them ideal for structures that require high strength but need not be heavy. They can be molded into any required shape that provides higher design flexibility. Moreover, it has a lower environmental impact (Lee and Jain 2009), unlike concrete and timber.

The load-carrying capacity of concrete structures is reduced by earthquakes or by the action of the freeze-thaw cycle, carbonation, chemical exposure, etc. Replacing a structural member is not a workable solution in an existing structure. The reduced load-carrying capacity can be improved by providing lateral confinement to the structural member using CFRP as external reinforcement (Taranu et al. 2008). Lateral confinement increases the axial compressive strength of structural members by developing a triaxial stress field. External confinement of structural members is a common way of retrofitting concrete columns (Lin and Liao 2004). Many empirical models have been developed on the compressive strength of the concrete confined using FRP. The performance of structural members depends on the material properties of CFRP including its type, strength, elastic modulus, its thickness, and number of layers used for lateral confinement of members (Dundar et al. 2015). The lateral confinement slows the collapse of concrete columns by reducing the rate of spalling of concrete cover (Ahmad et al. 2022). CFRP retrofitting of concrete members has been proven to be successful in improving ductility, stiffness, and load-carrying capacity. Hadi and Le (2014) prepared four groups of columns including unwrapped specimens and three groups of columns wrapped with CFRP with different fiber orientations. Each fiber orientation resulted in increased ductility and strength as compared to the unwrapped samples. The effect of CFRP on the performance of non-reinforced concrete was studied by Cao et al. (2019) and they discovered an increase in the ultimate load-carrying capacity of unreinforced members with an increase in the number of CFRP sheets wrapped around the member. The compressive behavior of ultra-high-performance concrete in circular columns confined by FRP was studied by Liao et al. (2021). An extensive database of 117 was prepared on ultra-high-performance concrete confined with FRP by Liao et al. (2022a) to estimate the stress and strain relationship under axial compression. Partial wrapping of FRP has been used by Liao et al. (2022b) instead of full wrapping to determine the performance of concrete columns confined with FRP spiral strips. This partial confinement resulted in increased ultimate axial stress and strain and improved the fire resistance.

Machine learning is becoming popular due to its ability to make accurate predictions about the performance of structural members. Researchers have used ANN and hybrid ANN for predicting the strength of materials. Previous research shows that the behavior of the FRP concrete can be predicted much better using machine learning models including ANNs, support vector machines (SVMs), and metaheuristic algorithms. The behavior of FRP-reinforced concrete is complex and depends not on the material only but also the skill of the worker. The behavior of FRP bars depends on multiple physical parameters such as concrete strength, type of fiber, bar diameter, fineness of sand, and shear lag effect (Başaran et al. 2022). This develops a very complex relationship between the parameters that change significantly with the type of FRP used (Başaran et al. 2022). Machine learning provides a reliable solution to this problem by developing a relationship between the parameters depending on the results from the experimental work. These relations are used by the models to predict the behavior of the FRP more properly than the empirical methods that were developed from small databases. Nguyen and Kim (2021) predicted the ultimate strength of rectangular columns using a hybrid machine-learning approach. Jahangir and Rezazadeh Eidgahee (2021) used a hybrid ANN based on an artificial bee colony for evaluating the bond strength of FRP concrete. Researchers have used hybrid ANN for the optimization of structural members. Jia et al. (2022) have used three metaheuristics-based models to determine the numerical performance of debonding strength in FRP composites. Metaheuristics-based ANN models were used to determine the effect of an alkaline concrete environment on the durability of GFRP bars by Khan et al. (2022), and degradation of GFRP tensile strength was determined using three different hybrid ANN models. A novel hybrid machine learning model of response surface model coupled with support vector regression was used by Keshtegar et al. (2021) to predict the ultimate condition of the concrete confined with FRP. Yan et al. (2017) have used a genetic algorithm-based ANN model to estimate the bond strength of GFRP bars in concrete.

Existing research is focused on either empirical models using a small experimental dataset that lacks all the combinations of the variables involved. In the present study, three metaheuristics algorithms including particle swarm optimization (PSO), grey wolf optimizer (GWO), and bat algorithm (BA) were combined with ANN to develop three hybrid ANN models, that were trained on 708 carbon fiber reinforced polymer (CFRP) confined concrete cylinders dataset to predict the strength of confined concrete. In this study, the results of the hybrid models are validated using the finite element analysis (FEA) employing ABAQUS for the experimental data of 18 CFRP-wrapped cylinders. All three hybrid ANN models are trained and tested utilizing an objective function of mean square error (MSE). Proposed hybrid models of PSO and GWO predicted the axial compressive strength of CFRP concrete with the highest accuracy than the empirical models. The high accuracy of the prediction models is useful in replacing the long and tedious experimental process and is also helpful for analyses and design of CFRP-confined members.

## 2  Confinement Mechanics

The FRP confinement enhances the strength and durability of the concrete (Mirmiran et al. 1998, Ahmad et al. 2020, Rodsin 2021, de Diego et al. 2022). It increases the lateral pressure on the concrete, which reduces its lateral expansion and improves axial strength. Various researchers have provided strength models for the FRP confined concrete (Mander et al. 1988). The confinement mechanics of FRP confined concrete has been defined using some common parameters including hoop rupture strain of fibers, strain ratio, confinement stress, and confinement stiffness ratio. Lim et al. (2016) used genetic programming to develop a model for the FRP-confined concrete to predict the ultimate conditions of the FRP concrete. They established expressions for estimating hoop rupture strain ($\varepsilon_{h,rup}$) of FRP against the maximum tensile strength of fibers ($\varepsilon_f$) using Equation 1.

$$\varepsilon_{h,rup} = \left(\frac{\varepsilon_f}{f_{co}^{'0.125}}\right) \tag{1}$$

Previous models developed by Richart et al. (1929) and Newman and Newman (1971) used lesser parameters for the geometrical and mechanical properties making them unreliable for future studies. The existing models provided a poor prediction of the test results for axial strain at compressive strength of confined concrete. The empirical model proposed by Lam and Teng (2003) provided a simple design-oriented stress-strain model for axial strain and compressive strength. The models of Lam and Teng (2003) were changed a little bit by the ACI and were added to the ACI 440.2R-08 design guidelines. The empirical strength model developed by Teng et al. (2009) presented the equations for the strain ratio and confinement stiffness ratio as given in Equations 2 and 3.

$$p_\epsilon = \frac{\epsilon_{h,rup}}{\epsilon_{co}} \tag{2}$$

$$p_k = \frac{2E_f t}{\left(\frac{f'_{co}}{\epsilon_{co}}\right)D} \tag{3}$$

where $p_\epsilon$ and $p_k$ are the strain ratio and confinement stiffness ratio respectively, $E_f$ is the elastic modulus of FRP sheets in the transverse direction, t is the thickness of sheets, $\epsilon_{co}$ is the axial strain of unconfined concrete and $f'_{co}$ is the compressive strength of the unconfined concrete. The relation of the maximum confinement stress based on the confinement mechanics of the CFRP sheets as illustrated in Figure 1 is as follows:

$$f'_{co} = \frac{2E_f \varepsilon_{h,rup} t}{D} \tag{4}$$

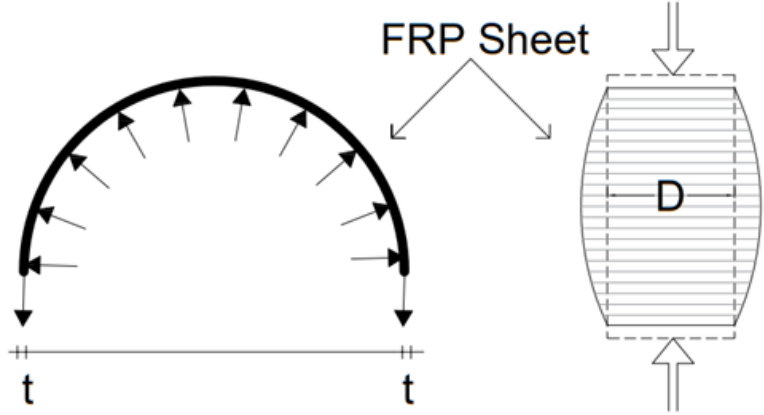

**Figure 1.** Confinement mechanics of CFRP.

## 3   Artificial Neural Network

A neural network is a machine-learning model inspired by the structure and function of the human brain. It is composed of interconnected artificial neurons that can be trained to recognize patterns and make predictions on input data. The basic building block of a neural network is the artificial neuron, which takes one or more inputs, performs a computation on them, and produces one output. A feed-forward neural network is the most common type of neural network, consisting of layers of artificial neurons connected in a directed acyclic graph, where the input is passed through the layers one at a time to the input layer, this input data is used to perform computations in the hidden layers and the output is produced by the final output layer. The computation performed by a neuron is a linear combination of its inputs, followed by a non-linear activation function. The linear combination of inputs and weights is represented mathematically as given in equation 5:

$$Z = Wx + b \tag{5}$$

where *x* is the input, *W* is the weights, and *b* is the bias. The activation function, *Z* produces the output of the neuron based on the input. Neural networks use different activation functions including sigmoid, ReLU, and tanh. The parameters of a neural network are the weights and biases of the connections between the neurons, which determine the computation performed by each neuron. The learning algorithms optimize these parameters to minimize the difference between the network's predicted outputs and the desired outputs, which is the process known as training. The weights are changed based on the difference between the target value and the predicted value (Basheer and Hajmeer 2000). In addition to the optimization algorithm, other factors can affect the performance of a neural network, such as the architecture of the network and the amount and quality of the training data (Chaturvedi 2008). Architecture refers to the number of layers, the number of neurons in each layer, and the connections between the layers, as displayed in **Error! Reference source not found.**. A deeper and wider architecture with more neurons and layers can increase the capacity of the network, but it also increases the risk of overfitting, where the network memorizes the training data instead of generalizing it to new data. On the other hand, a shallower and narrower architecture with fewer neurons and layers can decrease the capacity of the network, but it also reduces the risk of overfitting.

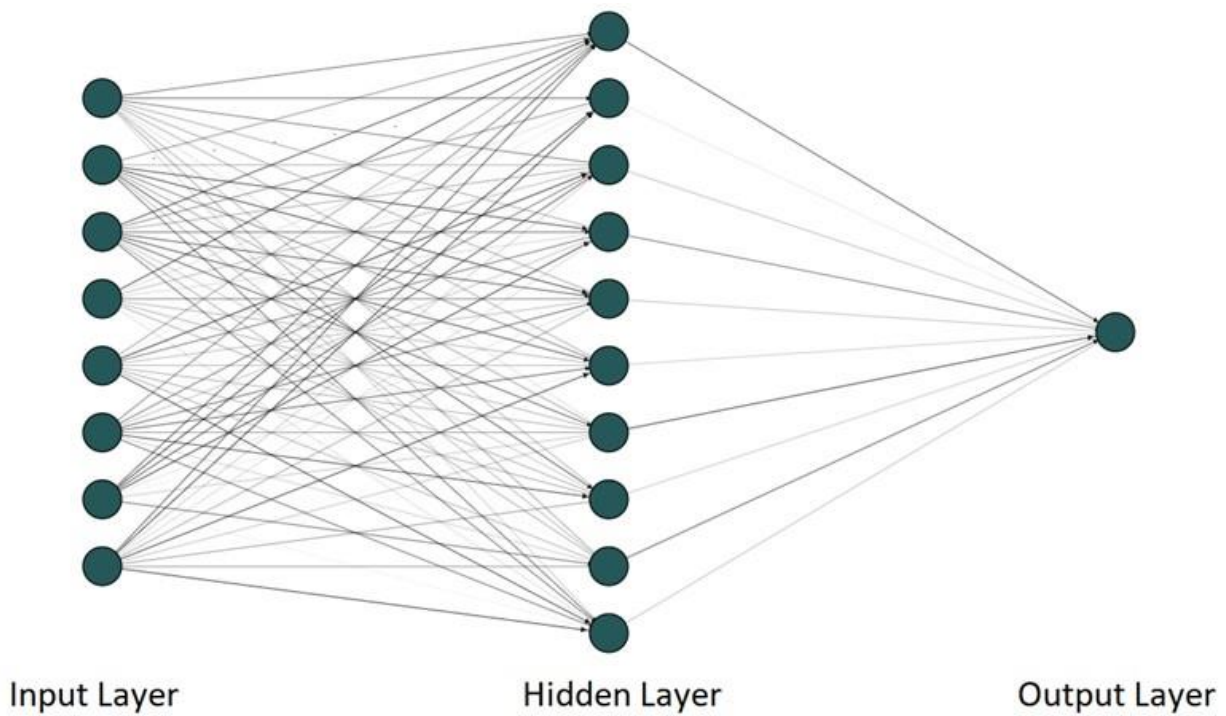

**Figure 2.** Neural network architecture.

A key concept in the architecture is the regularization of the model. This is done to avoid overfitting and improve the generalization capability of the network. There are various forms of regularization like dropout and early stopping. Finally, the amount (Ajiboye et al. 2015, Linjordet and Balog 2019, Bailly et al. 2022) and quality of the training data plays a critical role in the performance of a neural network. A larger and more diverse training set can increase the ability of the network to generalize to new data, but it also increases the computational cost of training. The quality of the data is also important, as noisy or biased data can impact the performance of the network (Gupta and Gupta 2019). Preprocessing the data and removing noise and outliers can improve the quality of the data and lead to better performance (Shanker et al. 1996). Training a neural network is a complex process that involves multiple parameters and techniques, such as the optimization algorithm, the architecture, and the amount and quality of the training data. The best approach is to experiment with different combinations of these factors to find the optimal settings for a particular problem and dataset.

Each type of neural network is suitable for a specific use case, and the choice of which one to use depends on the characteristics of the problem and the data. In addition to the various types of neural networks, several techniques can be used to improve the performance of neural networks, such as data preprocessing, hyperparameter tuning, and ensemble methods. Data preprocessing is the process of preparing the data for use in a neural network. This includes tasks such as normalizing or scaling the data, removing outliers or noise, and transforming the data into a suitable format. Finally, it's important to note that neural networks can be computationally intensive, especially for large and complex problems, so it's important to consider the computational resources and constraints of the problem. This might include using specialized hardware, such as graphical processing units (GPUs), or using cloud-based services to train the networks.

## 4  Hybrid ANN

A hybrid ANN is a type of neural network that combines the strengths of multiple models or algorithms to improve its performance. This can be achieved by integrating other models or algorithms with an ANN, to leverage their strengths and overcome the limitations of a traditional ANN. Hybrid ANNs can be developed in different ways, depending on the specific problem and the models or algorithms that are being combined. Some examples of developing a hybrid ANN include combining ANNs with other machine learning models, such as SVMs, decision trees, or k-nearest neighbors (k-NN) (Chu et al. 2021). A hybrid ANN could be developed by combining an ANN with SVM, where ANN is used to extract features from the input data, and the SVM is used to classify the data. ANNs can be combined with evolutionary algorithms, such as genetic algorithms (GAs) (Soleimani et al. 2018) or PSO (Shariati et al. 2019). A hybrid ANN could be developed by training an ANN using a GA, where the GA is used to optimize the weights of ANN. ANNs can be combined with rule-based systems, such as fuzzy logic or expert systems (Sivaneasan et al. 2017). A hybrid ANN could be developed by integrating a fuzzy logic system with an ANN, where the fuzzy logic system is used to handle the imprecision and uncertainty of the input data, and ANN is used to perform the computation.

The main advantage of a hybrid ANN is that it can achieve better performance than a traditional ANN, due to the combination of multiple models or algorithms. For example, a hybrid ANN that combines an ANN with a metaheuristic algorithm can achieve better performance than an ANN trained using a traditional optimization algorithm because the metaheuristic algorithm can explore the search space more efficiently (Kaveh 2021) but this is not always the case. Similarly, a hybrid ANN that combines with rule-based systems can handle imprecision and uncertainty of data more efficiently. Another advantage is that hybrid ANNs are more robust and less prone to overfitting than traditional ANNs (Kaveh 2021), as combining multiple models can help to reduce the risk of overfitting by averaging the errors of individual models.

### 4.1  Particle Swarm Optimization (PSO)

PSO is a population-based metaheuristic algorithm that is inspired by the social behavior of bird flocks or fish schools (Poli et al. 2007). It is used to find the global optimum of a given objective function. The optimization ability of a flock of birds was studied by Kennedy and Eberhart who developed PSO in 1995. PSO has several variants including constriction PSO (CPSO), inertia weight PSO (IWPSO), fully informed PSO (FIPSO) (Mendes et al. 2004), and adaptive PSO (APSO). In a hybrid ANN, PSO is used to optimize the weights of ANN. It is used to find the best set of weights that minimize the error between the predicted output and the actual output. This process is done by iteratively adjusting the weights based on the PSO algorithm's results. The PSO algorithm is used to find the best set of weights for ANN, and ANN is used to predict the output based on the input data and the optimized weights. The PSO algorithm has several parameters that can be adjusted to control the behavior of the optimization process including population size, the maximum and minimum values of the search space, inertia weight, cognitive weight, and social weight, as depicted in **Error! Reference source not found.**.

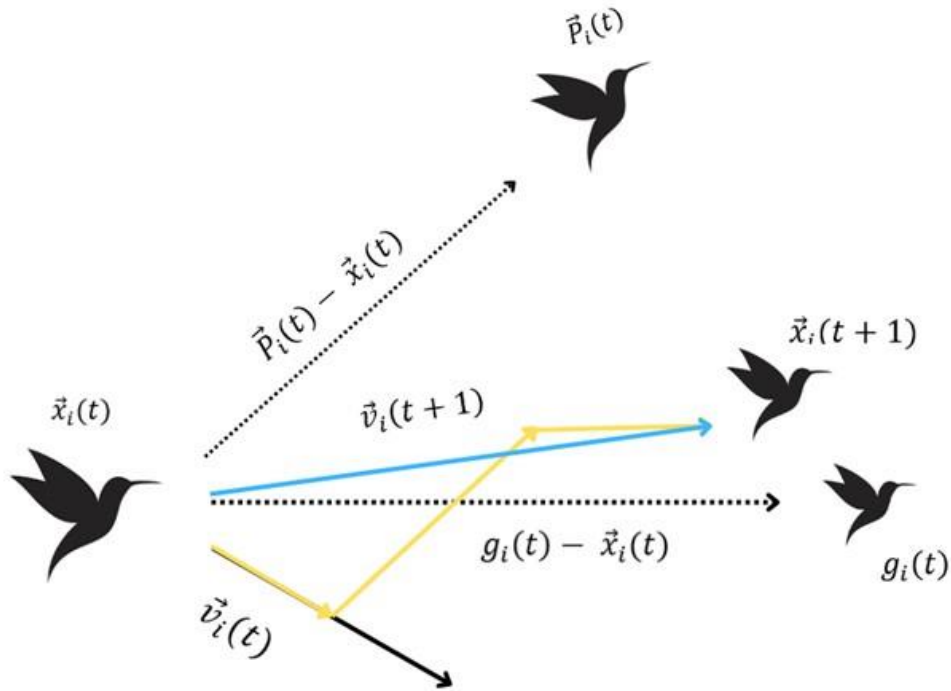

**Figure 3.** Velocity and position adjustment in PSO.

The population size determines the size of the swarm or the number of particles that are used for optimization. The chances of finding a global optimum increase with the increase in the swarm size but it becomes computationally expensive as the number of particles increases. The optimum value of the swarm size is dependent on the problem and therefore it may vary. The maximum and minimum values of the search space determine the range of values that the particles can take relevant to the research problem being solved. The working of the PSO is presented in a flowchart in **Error! Reference source not found.**.

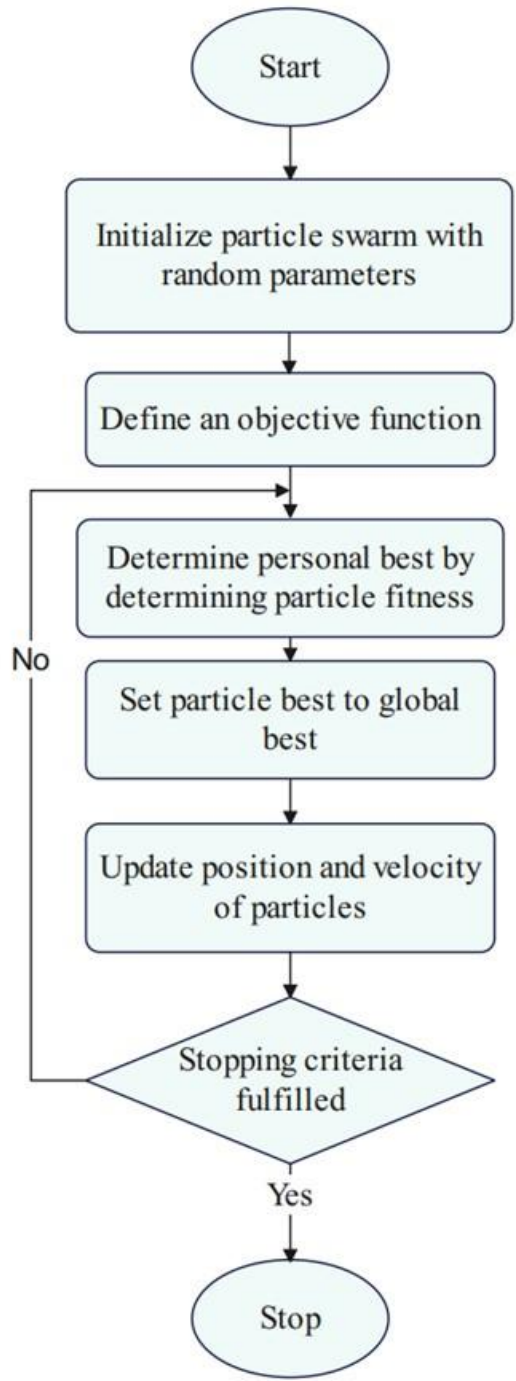

**Figure 4.** PSO flowchart.

The exploration and exploitation behavior of the algorithm is controlled by the inertia weight, cognitive weight, and social weight. The inertia weight determines the influence of the previous velocity of the particle on its current movement. A higher value means that the particle will be more biased towards the previous movement and a lower value of inertia weights will cause the particle to be more responsive towards the other particles in the swarm and global best position. The cognitive weight controls the tendency of the particle to move towards its own personal best position. A higher value means that the previous personal best position of the particle will have a greater impact on the current movement. The social weight controls the tendency of the particle to move towards the global best position of the swarm. Increasing inertia weight will increase the tendency of the particle to move in the same direction as it did in the previous iteration. Increasing cognitive weight will increase the tendency of the particle to move towards its own personal best position, which can make the algorithm more aggressive in exploring the search space. Increasing social weight will increase the tendency of the particle to move towards the global best position of the swarm. Increasing the value of all these three weights can make the algorithm converge faster but also increases the chances of getting stuck in a local optimum.

### 4.2 Grey Wolf Optimizer (GWO)

GWO is a metaheuristic algorithm that was inspired by the hunting behavior of grey wolves (Mirjalili et al. 2014). It is used to solve optimization problems and find the global optimum of a given objective function. The GWO algorithm has several parameters that can be adjusted to control the behavior of the optimization process. These include the population size, the maximum and minimum values of the search space, and the alpha, beta, and delta coefficients. Each parametric value affects the working of the optimization algorithm. The population size determines the number of solutions that are considered at each iteration of the optimization process. The maximum and minimum values of the search space determine the range of values that the solutions can take. These values should be set based on the specific problem being solved.

The alpha, beta, and delta coefficients are used to control the exploration and exploitation behavior of the algorithm. The alpha coefficient controls the step size of the search, the beta controls the direction of the search, and the delta controls the convergence rate of the algorithm. Increasing alpha will increase the step size of the search, which can speed up the optimization process but also increases the chances of missing the global optimum. Increasing beta will increase the direction of the search, which can make the algorithm more aggressive in exploring the search space, but also increases the chances of the algorithm getting stuck in a local optimum. Increasing the delta will increase the convergence rate of the algorithm, which can make the algorithm converge faster, but also increases the chances of missing the global optimum.

### 4.3 Bat Algorithm (BA)

BA is a population-based optimization algorithm that is inspired by the echolocation behavior of bats (Yang and Gandomi 2012). Like the other two algorithms discussed above, it is also used to find the global optimum of a given objective function. The optimization process is controlled by the parameters involved including the population size, the maximum and minimum values of the search space, and the frequency range, loudness, and pulse emission rate (alpha). The working of BA from the initialization of the swarm to the end criterion is indicated in a flowchart in **Error! Reference source not found.**. The population size determines the number of bats that are considered at each iteration of the optimization process. The maximum and minimum values of the search space determine the range of values that the solutions can take. These values should be set based on the specific problem being solved. The frequency range is used to control the exploration and exploitation behavior of the algorithm. Increasing the frequency range will increase the chances of finding the global optimum, but also increases the computational time required. The loudness controls the exploration and exploitation behavior of the algorithm. Increasing loudness will increase the exploration and exploitation behavior of the algorithm which can make the algorithm more aggressive in exploring the search space, but also increases the chances of the algorithm getting stuck in a local optimum. The pulse emission rate (alpha) controls the convergence rate of the algorithm. Increasing the pulse emission rate (alpha) will increase the convergence rate of the algorithm which can make the algorithm converge faster, but also increases the chances of missing the global optimum.

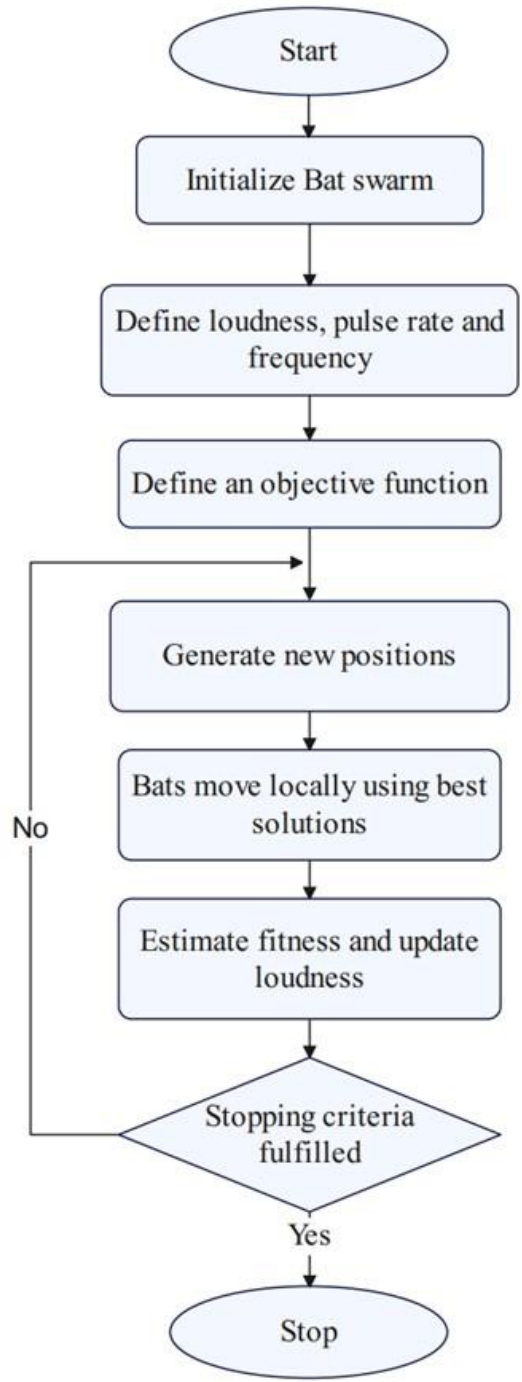

**Figure 5.** BA flowchart.

## 5 Database for CFRP Confined Cylinders

To predict the axial compressive strength of concrete members confined with FRP, a dataset of 708 samples was developed from the previous research publications. The common variables of the data are chosen for the training and testing of ANN and hybrid ANN. Seven common features in the dataset are used as the input variables including the geometrical properties such as the diameter of the cylinder ($D$), the height of the cylinder ($H$), total thickness ($n_t$), and elastic modulus of the FRP ($E_s$), and compressive strength of the unconfined concrete ($f'_{co}$), unconfined concrete strain ($\varepsilon_{co}$), confined concrete strain ($\varepsilon_{cc}$). The confined compressive strength of the CFRP concrete is used as the target variable for the prediction models. This data is further divided into training and testing data in 75% and 25% proportions.

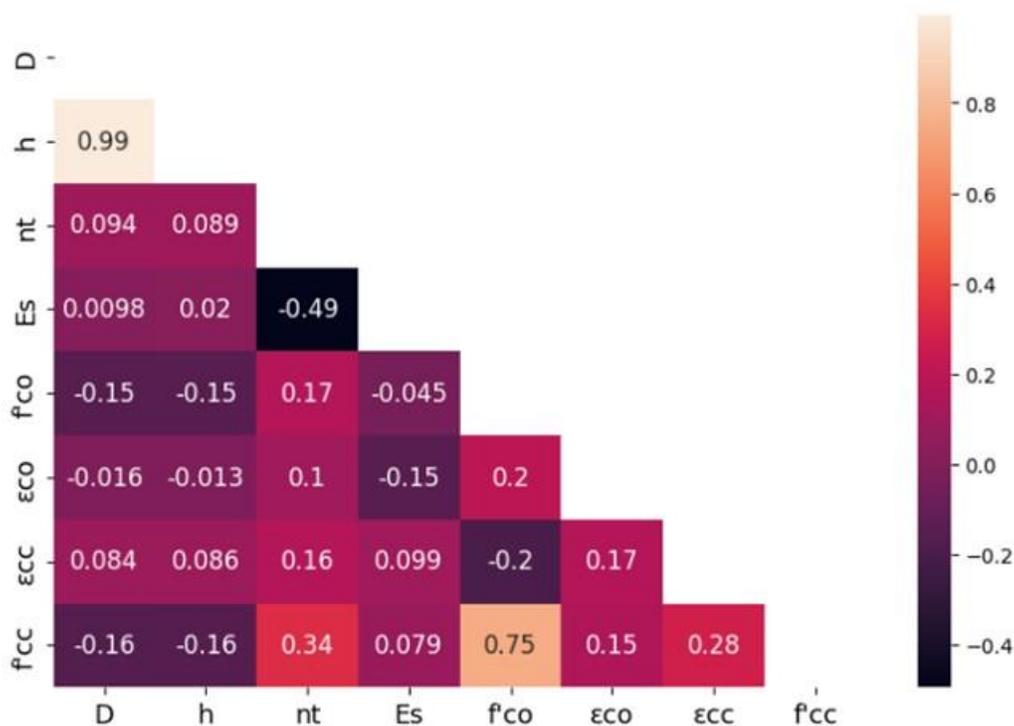

**Figure 6.** Correlation Heatmap for dataset.

A statistical description of the data is provided in **Error! Reference source not found.** and the correlation of dataset parameters is reported in Figure 6, which shows the influence of the variable on each other's value and most importantly on the compressive strength of the CFRP confined cylinders. The heatmap reveals color variation with the change in the influence of parameters on each other values. The darker region denotes a lesser correlation, and the light region signifies a high correlation between the parameters. The main parameter of concern is the axial compressive strength and the last row of the heatmap elaborates on the influence of each parameter on its value. The unconfined compressive strength and the thickness of CFRP sheets have the most impact on the confined compressive strength of the concrete cylinders.

**Table 1.** Statistical description of dataset

| Parameter | D | h | $n_t$ | $E_s$ | $f'_{co}$ | $\varepsilon_{co}$ | $\varepsilon_{cc}$ | $f'_{cc}$ |
|---|---|---|---|---|---|---|---|---|
| | (mm) | (mm) | (mm) | (GPa) | (MPa) | (%) | (%) | (MPa) |
| Min | 51 | 102 | 0.09 | 10 | 12.41 | 0.1676 | 0.083 | 18.5 |
| Max | 406 | 812 | 5.9 | 663 | 188.2 | 1.53 | 4.62 | 302.2 |
| Diff | 355 | 710 | 5.81 | 653 | 175.79 | 1.3624 | 4.537 | 283.7 |
| Mean | 153.34 | 306.45 | 0.89 | 174.68 | 42.48 | 0.27 | 1.54 | 76.25 |
| Median | 152 | 304 | 0.47 | 219 | 37.7 | 0.24 | 1.35 | 67.91 |
| St. Dev. | 43.16 | 85.67 | 1.05 | 118.8 | 22.39 | 0.14 | 0.85 | 35.01 |
| Cov | 0.29 | 0.28 | 1.18 | 0.69 | 0.53 | 0.52 | 0.56 | 0.46 |

The database contains eight variables of different ranges, which reduces the effectiveness of the ANN models. Some variables like the depth of cylinders have very large values in multiples of 100, while the strain values are less than 1. This large variation in data creates troubles for the prediction models during the training and testing of the data. Therefore, the data is converted between suitable upper and lower limits that convert all the variables within a single range and increase the efficiency of the ANN model. The database is normalized between 0.1 and 0.9 rather than 0 and 1, for all the parameters of the CFRP-confined concrete cylinders using Equation 6.

$$X = \left(\frac{0.8}{x_{max} - x_{min}}\right)x + \left(0.9 - \left(\frac{0.8}{x_{max} - x_{min}}\right)x_{max}\right) \tag{6}$$

where $x_{max}$ and $x_{min}$ are the maximum and minimum values of the parameters of the dataset.

## 6 Model Development

This research uses four different prediction models including one model of feed-forward neural network (FNN) and three hybrid neural networks based on the metaheuristic algorithms of GWO, PSO, and BA combined with the FNN. In the FNN model, data flows in only a forward direction through the network. It contains input, hidden, and output layers that process the data through them via a combination of the weights and biases that are continuously updated in each iteration using an optimization algorithm. The neural networks can develop a pattern in the data to perform predictions, but they can face several problems during training and deployment stages including overfitting, underfitting, and getting trapped in local minima. Metaheuristics are used to optimize the architecture and topology of ANN, which can lead to a more complex and expressive model that can better fit the data to overcome the problem of underfitting (Khan et al. 2021). These algorithms can be used to optimize the weights and biases of ANN (Ojha et al. 2017) in a way that balances the trade-off between fitting the training data and generalizing it to new, unseen data to overcome the problem of overfitting. They can avoid getting trapped in local minima by exploring and exploiting the entire search space to achieve the global minimum.

However, the hybrid ANN models don't need to always perform better than the ANN model. The metaheuristic algorithms of PSO, GWO, and BA have their own set of parameters that have unique optimum values depending on the database and research problem. These are run on the database multiple times by varying their parametric values to optimize their prediction process and reduce the computational cost. The optimum values of the parameters for each of the metaheuristic models are observed in **Error! Reference source not found.**.

**Table 2.** Hybrid Model Parameters

| Sr. No | Model | Population | Iterations | Parametric Values | Neurons | Objective Function |
|---|---|---|---|---|---|---|

| # | Algorithm | | | Inputs | | Objective |
|---|---|---|---|---|---|---|
| 1. | PSO | 70 | 900 | D, h, nt, $E_f$, $f'_{co}$, $f'_{cc}$, $\varepsilon_{cc}$, $\varepsilon_{co}$ | 50 | MSE |
| 2. | GWO | 75 | 900 | D, h, nt, $E_f$, $f'_{co}$, $f'_{cc}$, $\varepsilon_{cc}$, $\varepsilon_{co}$ | 50 | MSE |
| 3. | BA | 80 | 900 | D, h, nt, $E_f$, $f'_{co}$, $f'_{cc}$, $\varepsilon_{cc}$, $\varepsilon_{co}$ | 50 | MSE |

All three algorithms use the objective function of MSE and the model resulting in the maximum value of R and lowest MSE value is the best model. The optimum value of the parameters of the algorithms is also decided based on these results and the computational time. The input parameters used for the FNN and the hybrid neural networks of PSO, GWO, and BA are given in **Error! Reference source not found.** with their statistical description. These three algorithms are compared with each other based on the values of the error metrics of *MSE*, *MAE*, and the correlation factor ($R^2$) that are expressed in the following equations.

$$MSE = \frac{\sum_{i=1}^{n}(Y - \bar{Y})^2}{n} \quad (7)$$

$$MAE = \frac{\sum_{i=1}^{n}|Y - \bar{Y}|}{n} \quad (8)$$

$$R^2 = \left(\frac{n(\sum xy) - (\sum x)(\sum y)}{\sqrt{[n\sum x^2 - (\sum x)^2][n\sum y^2 - (\sum y)^2]}}\right)^2 \quad (9)$$

where $Y$ and $\bar{Y}$ are the experimental and predicted values respectively. The accuracy of the prediction model is determined by the values of the error metrics of *MSE* and *MAE* and $R^2$. The prediction model with the lowest values of *MSE*, *MAE*, and the highest value of $R^2$ is the best. There is still another factor which is the biases of the prediction model towards either underestimation or overestimation of the predicted values, this is determined by the plots of predicted values.

## 6.1 Comparison of Empirical and Hybrid ANN Models

This research work is based on two empirical models, three hybrid ANN models, and one ANN model to predict the axial compressive strength of CFRP concrete cylinders using a database of 708 samples. The models proposed by Lam and Teng (2003) and Miyauchi et al. (1997) are used in the present study after careful analysis of the other empirical models. The models proposed by Lam and Teng (2003) and Miyauchi et al. (1997) were tested against the experimental values and their accuracy was higher than other empirical models. The models of Lam and Teng and Miyacuhi et al. (1997) are given in Eq 10 and 11 respectively. The accuracy and error metrics of these models are represented in **Error! Reference source not found.** and **Error! Reference source not found.**, having $R^2 = 97.5$, *MSE* = 1.86% for Lam and Teng (2003) model and $R^2 = 97.4\%$, MSE = 2.01% for the Miyauchi et al. (1997) model. The author also proposed a nonlinear empirical model in the previous research (Ahmad et al. 2020) for the calculation of strength of confined concrete as given by Eq 12.

$$\frac{f'_{cc}}{f'_{co}} = 1 + 3.3\frac{f_l}{f'_{co}} \quad (10)$$

$$\frac{f'_{cc}}{f'_{co}} = 1 + 3.485\frac{f_l}{f'_{co}} \quad (11)$$

$$\frac{f_{cc}}{f_{co}} = 1 + k(\frac{f_l}{f_{co}})^n \quad (12)$$

The working of the hybrid ANN was optimized by tuning the parameters of each algorithm. This optimization was done by using *MSE* as an objective function. The preliminary evaluation of the models was done based on the error metrics of *MSE*, *MAE*, and $R^2$. The optimization is done by reducing the difference between the actual experimental values and the predicted values. The model with the least value of *MSE* and the highest value of *MAE* is the best model of all. Three different metaheuristic algorithms were used including PSO, GWO, and BA.

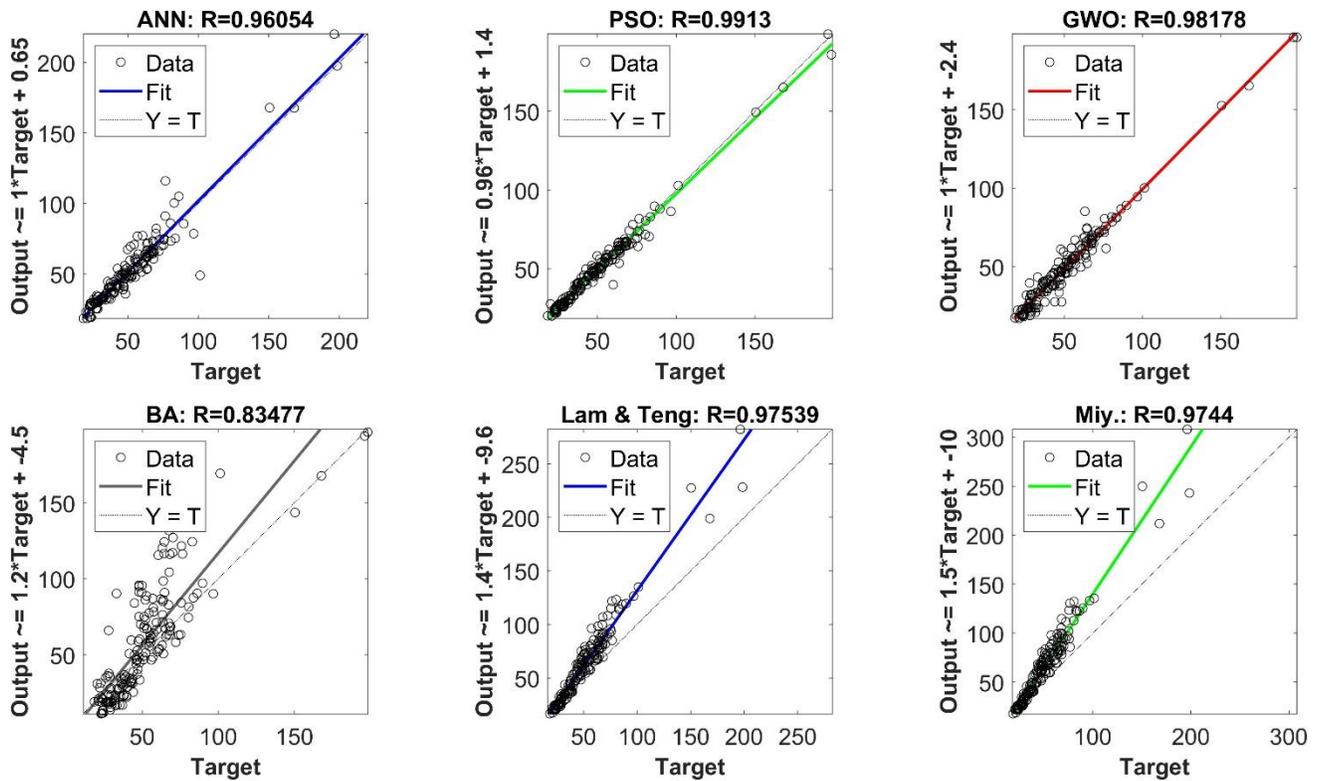

**Figure 7.** Accuracy of prediction models and empirical models.

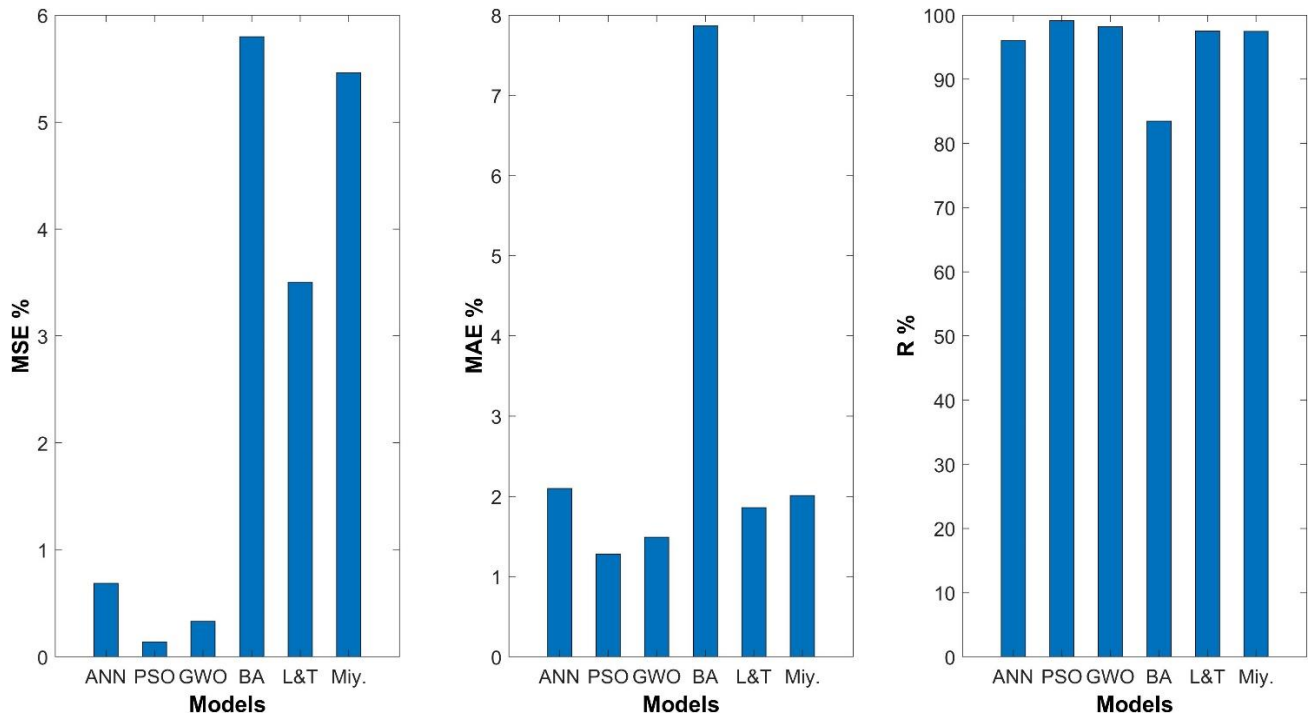

**Figure 8.** Error metrics of prediction models and empirical models.

The *MSE* and $R^2$ values have demonstrated that PSO and GWO-based ANN models provided better results for the prediction of axial compressive strength of the CFRP confined cylinder. The hybrid models were trained on different objective functions initially to determine the best optimization function for the current research. It has been observed that the working of the hybrid models changed significantly by varying the objective functions, keeping all other parameters of the model constant. The hybrid models of PSO and GWO provided the most accurate predictions based on the value of the error metrics of *MSE*, *MAE*, and $R^2$ value. The PSO model predicted the axial compressive strength of CFRP-confined cylinders with an accuracy of 99.13% followed by the GWO model with an accuracy of 98.17% and ANN with an accuracy of 96.05%. The error metrics for the PSO model were the lowest with *MSE* of 0.1414% and *MAE* of 1.28% while the GWO model has *MSE* of 0.33% and *MAE* of 1.49%. The ANN model also gave good results comparable to that of the GWO model. The predicted results from ANN, PSO, and GWO show that the predicted values of axial compressive strength are distributed uniformly around the best-fit line. While the empirical methods from Lam and Teng (2003) and Miyauchi et al. (1997) gave an overall good accuracy but they have a bias towards overestimation of the results. The values of axial compressive strength are found to be often more than the original strength values. Other than the prediction models, the two empirical models of Lam and Teng (2003) and Miyauchi et al. (1997) predicted the axial compressive strength with an accuracy of 97.53% and 97.44% respectively.

The predicted results of the testing data were analyzed against the target values for the axial compressive strength for each of the four models individually. The testing data for the models do not follow any particular pattern and were chosen randomly from the overall database. The predictions from the PSO, GWO, and ANN followed the rapidly changing data points of the testing data perfectly, as displayed in Figures 9-12**.**

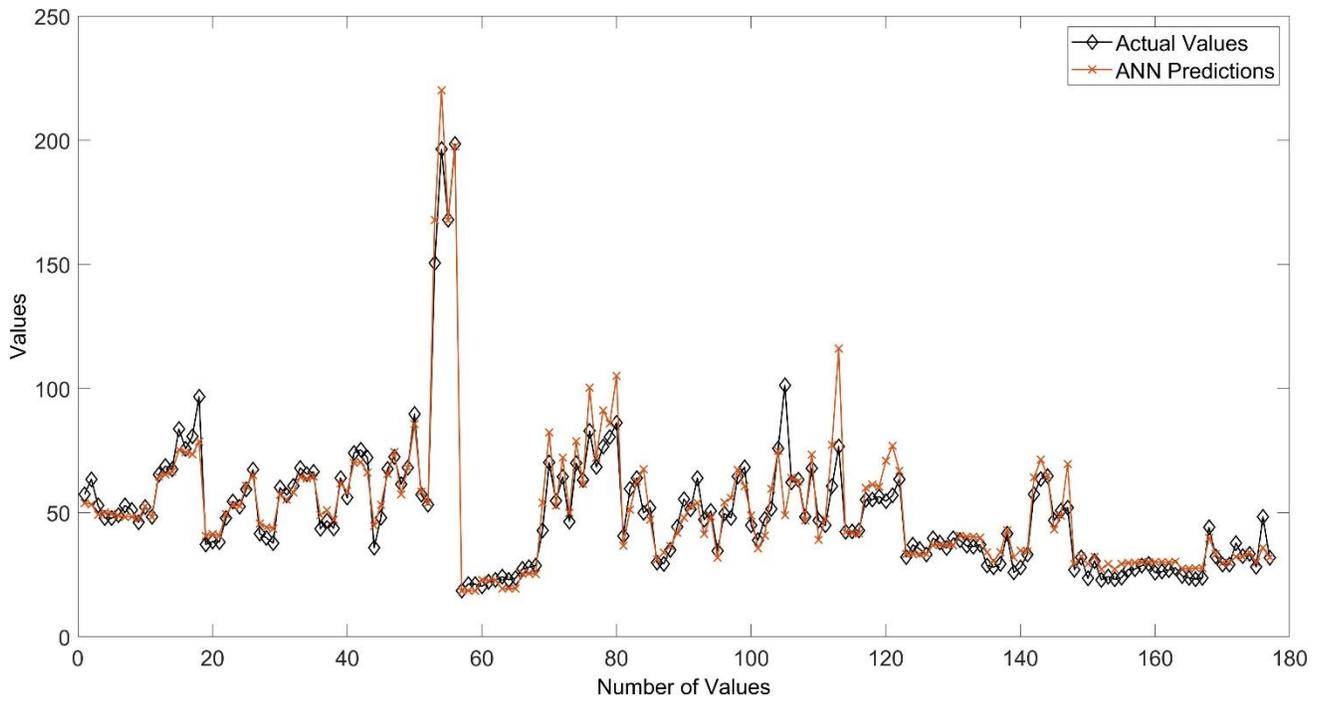

**Figure 9.** ANN predictions of axial compressive strength.

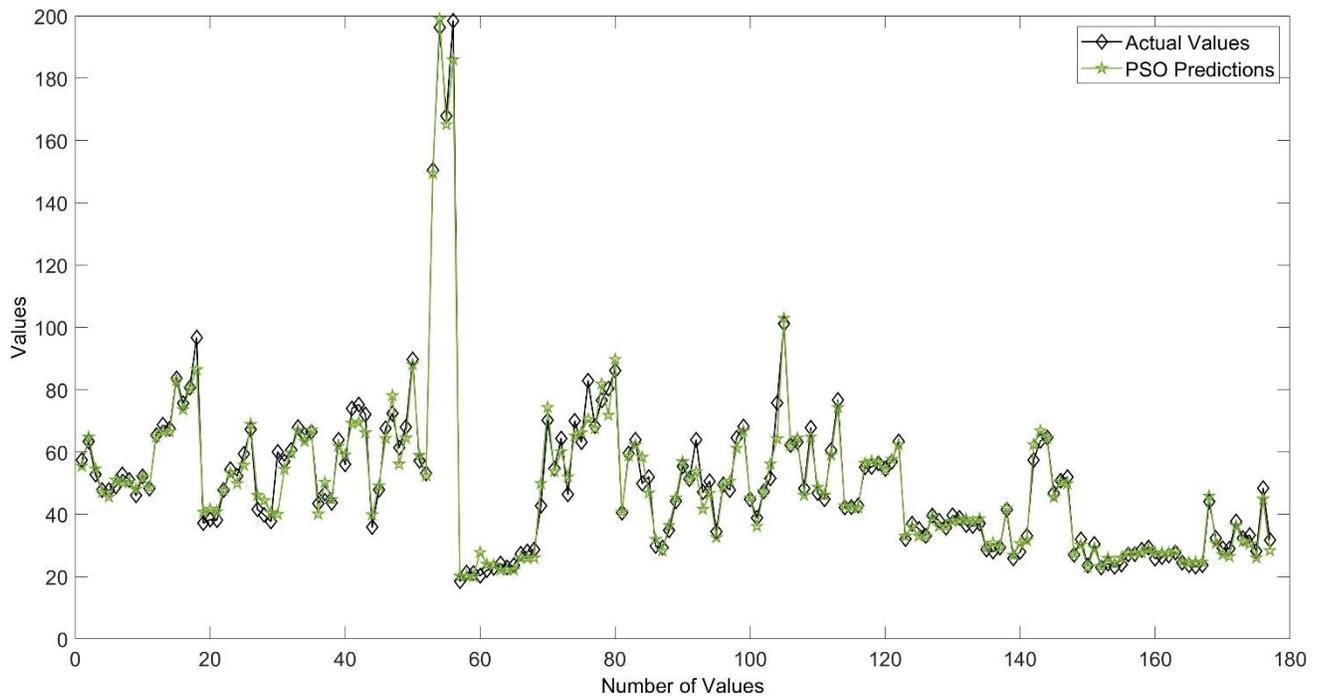

**Figure 10.** PSO predictions of axial compressive strength.

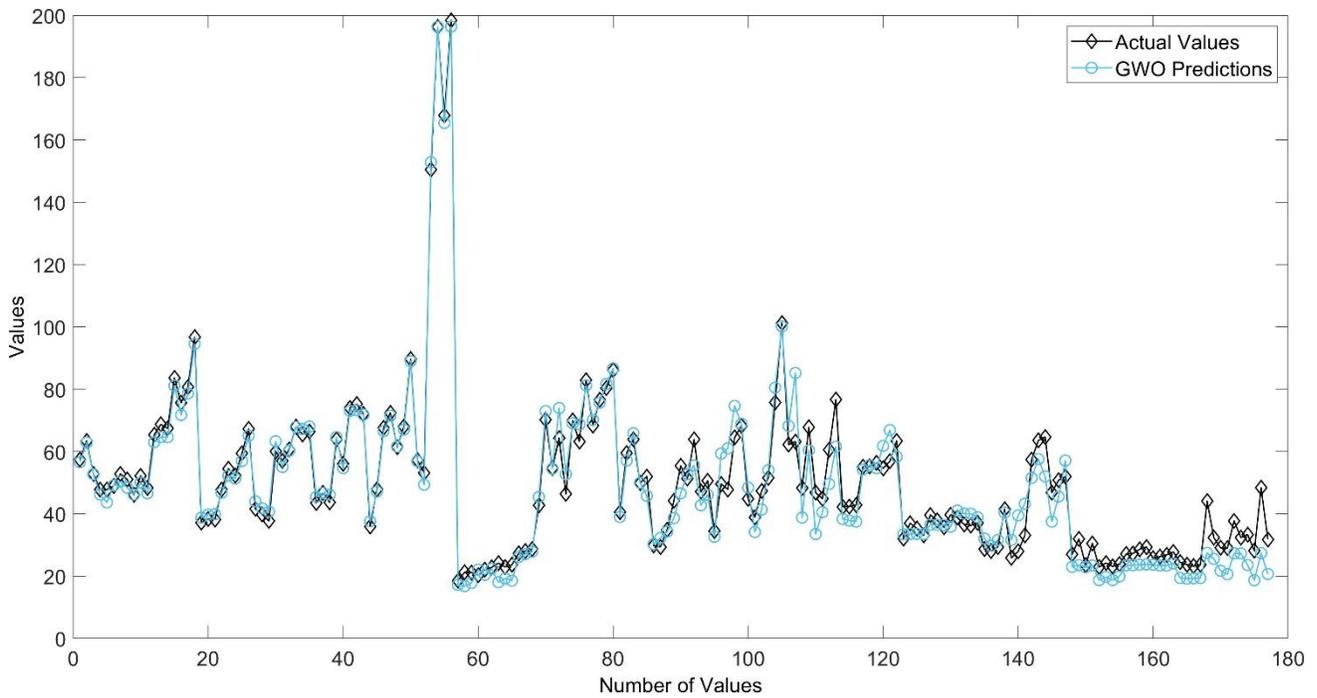

**Figure 11.** GWO predictions of axial compressive strength.

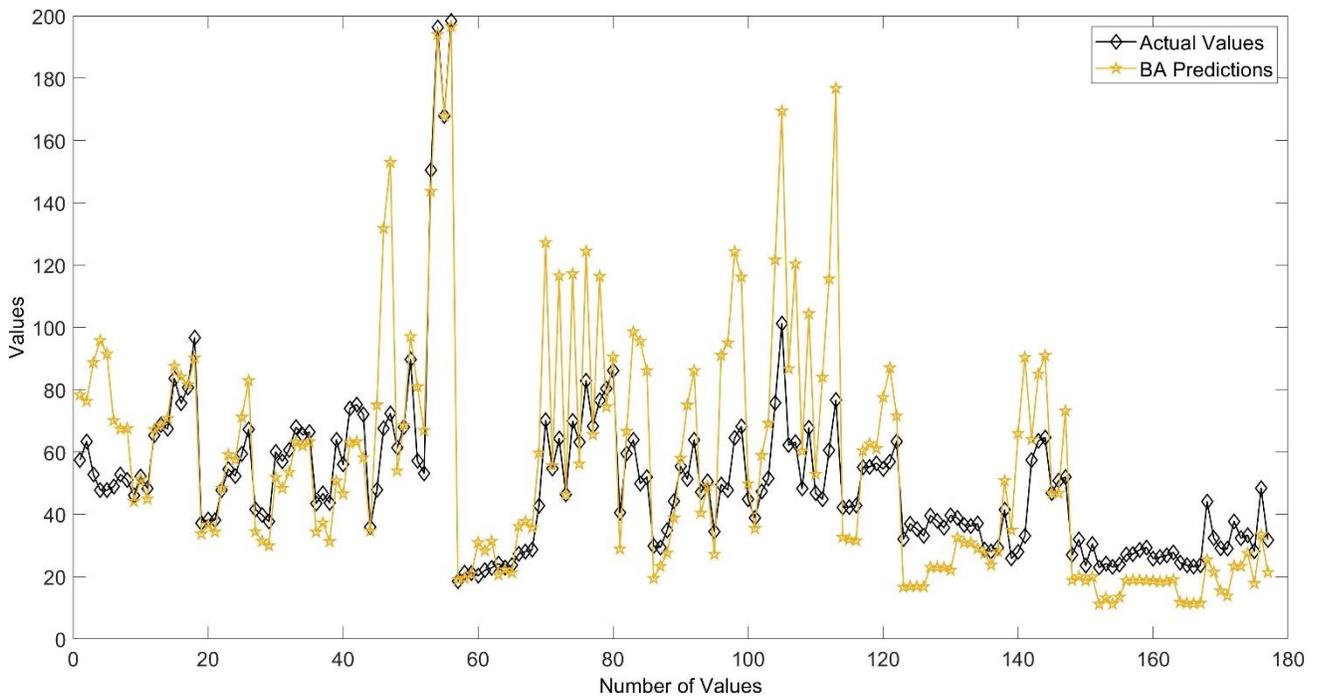

**Figure 12.** BA predictions of axial compressive strength.

The accurate predictions for this randomly and rapidly changing data uncover that the PSO, GWO, and ANN can be used as a substitute for empirical methods and experimental testing, making the overall process quick and economical. These predictions also reveal that not all metaheuristics-based hybrid ANN models perform accurate predictions. BA model predictions are far from the actual values of the axial compressive strength of CFRP-confined cylinders, as depicted in Figure 12. The normal distribution of the ratio of confined to unconfined strength for the experimental values and model predictions for the database are presented in Figure 13, and the distribution of confined to unconfined strength ratio of the database is indicated in Figure 14.

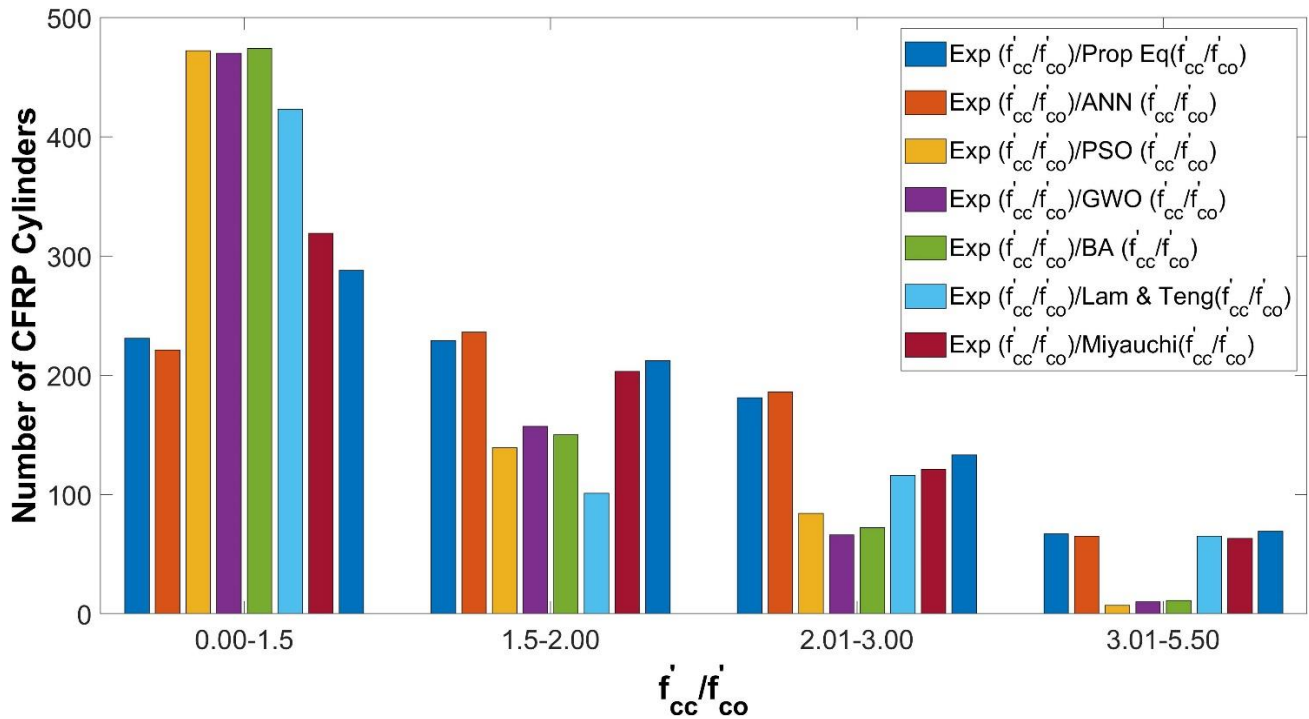

**Figure 13.** Distribution of confined to unconfined strength ratio of CFRP cylinders.

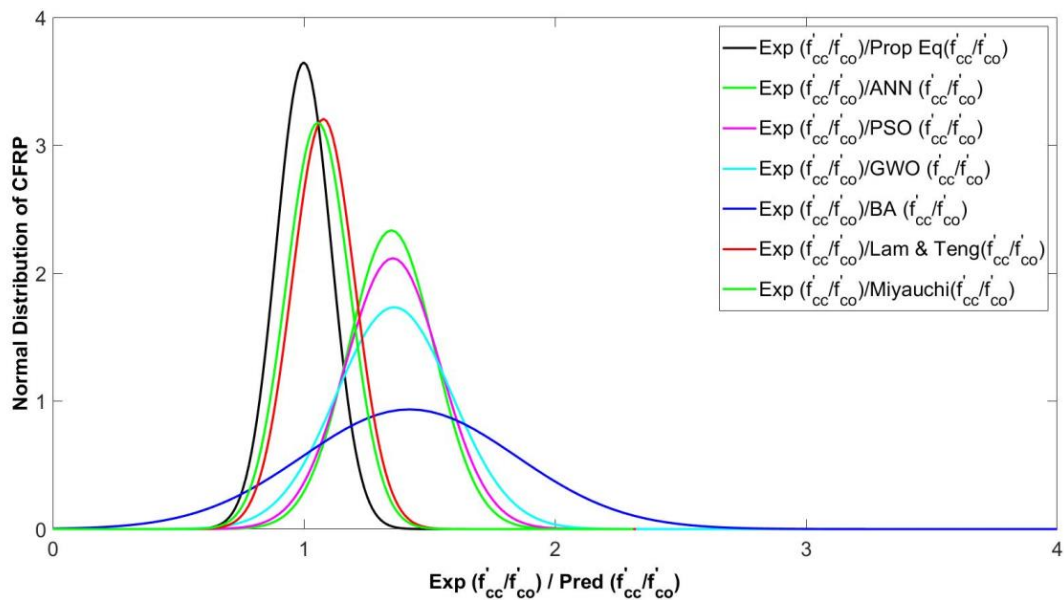

**Figure 14.** Normal distribution of $(f'_{cc}/f'_{co\,exp})/(f'_{cc}/f'_{co\,pred})$ for CFRP-wrapped cylinders.

## 6.2 Experimental Validation of Data

Research in concrete technology is focused on improving the mechanical properties of concrete and making the process economical. For this purpose, several supplementary cementitious materials, substitutes for aggregates, and even seawater as a replacement for freshwater have been used to determine the impact of changes in constituents on the concrete. The variation in materials significantly changes the properties of concrete and this variation shows specific patterns with different replacement proportions. But this can be accomplished during the casting of concrete only. It is not always a workable solution to increasing the size of the structural members if any deficiency is found after construction or if their capacity reduces during their design life due to natural hazards. Some problems including reinforcement corrosion, overloading, cracking, and aging of concrete may require more reinforcement that cannot be added once a structural member is built. These requirements can be fulfilled by wrapping FRP sheets around the structural members which do not increase the dimensions of the member but increase their load-carrying capacity.

The present research uses the author's previous experimental work on the 18 concrete cylinders. These cylinders used unidirectional wrapped CFRP sheets having a 4100 MPa nominal tensile strength and 231 GPa of nominal tensile elastic modulus that used 2-part epoxy impregnation resin of Sikadur-330 for the bonding of CFRP sheet with cylinders. It has a tensile strength of 30 MPa, elongation of 0.9%, and tensile elastic modulus of 4500 MPa. For the experimental investigation, a total of 18 concrete cylinders of standard size of 150 mm diameter and 300 mm height is cast initially. Nine cylinders have a compressive strength of 12.5 MPa and the other nines have 16.5 MPa compressive strength. In each group, three more distributions are made based on the number of CFRP sheets layer wrapped around them. In both groups of 9 cylinders, 3 cylinders are wrapped with 2 layers of CFRP, 3 cylinders with 1 layer of CFRP, and 3 cylinders are not wrapped with any CFRP layer. This division is done to determine the variation of the ductility and strength of concrete with varying numbers of CFRP layers around the concrete cylinders. The uniaxial test is performed to determine the compressive strength of all samples and it was observed that CFRP sheets increased the ductility and strength of concrete cylinders.

## 6.3  Finite Element Analysis

The finite element results are also taken from the author's previous work (Ahmad et al. 2020) where a concrete cylinder with one CFRP sheet wrapped around it was used as a numerical model for finite element modeling in ABAQUS. The behavior of the concrete and CFRP is different that requires behavior simulation models unique to the type of material. There are various models for simulating the behavior of concrete in ABAQUS including the concrete tension stiffening (CTS), continuum damage mechanics (CDM) model, extended finite element method (XFEM) model, and concrete damaged plasticity (CDP) model. The author's previous work used CDP which accounts for the damage and plastic deformation of the concrete. The CDP model includes parameters such as tensile strength, compressive strength, and fracture energy. The CDP model was used for the concrete and Hashin Damage Model for the FRP sheets.

The CDP model requires input for three important parameters of concrete to simulate the tensile, compressive, and plastic behavior. The tensile behavior of concrete was defined using the modified tension stiffening model and for compressive behavior, Eurocode 2 was used. The average compressive strain ($\varepsilon_{c1}$) and ultimate compressive strain ($\varepsilon_{cu1}$) were calculated using the following equations:

$$\varepsilon_{c1} = 0.0014[2 - e^{-0.024 f_{cm}} - e^{-0.140 f_{cm}}] \tag{13}$$

$$\varepsilon_{cu1} = 0.004 - 0.0011[1 - e^{-0.0215 f_{cm}}] \tag{14}$$

For calibration purposes, one concrete cylinder was chosen to identify the best approximation of each parameter on the stress-strain behavior of concrete. For this purpose, a concrete cylinder of 12.5 MPa compressive strength was calibrated by achieving a close agreement with the experimental results, then this model was calibrated for CFRP-wrapped samples. The geometrical parameters of the dilation angle, viscosity parameter, and mesh size of concrete were used for calibration of the finite element model (FEM) of concrete.

The dilation angle is calibrated for a close agreement between the FEA results and the stress-strain results of the experiment. Mesh size is an important parameter that significantly affects the analysis results. If the mesh size is too coarse, it results in strain localization because the energy is localized in a limited number of elements for that particular region of the model. This results in an overestimation of stresses and strains in the localized region. But a too-fine mesh size increases the computational cost of the simulation. It was observed that the stress-strain behavior of cylinders was not affected much by the variation in shape factor, stress ratio, and eccentricity, so the default values were used for these parameters. The effect of the element types of concrete was checked by studying all of the element libraries for 3D stress elements of concrete.

The Hashin Damage model was used for the simulation of the CFRP sheets wrapped around the concrete cylinders. The behavior of the CFRP in the ABAQUS can be described in two phases according to the actual behavior in the elastic and plastic phases. For the elastic stage, Poisson's ratio and elastic modulus are defined and for the plastic stage, the Hashin failure model is used. Hashin's theory for the damage prediction of the CFRPs is simulated on four damage mechanisms including compression and tension of both fiber and the matrix. The compressive stress and strains of the unconfined concrete cylinders and confined concrete cylinders are illustrated in Figure 15. The experimental results depicted that an increase in the CFRP layers increases the ductility and strength of the concrete cylinders. The authors' previous work indicated that the strength of the confined concrete cylinders increased from 69.42% to 87.2% for the 12.5 MPa strength concrete and the strength of confined concrete increased from 60.82% to 103.2% for 16.5 MPa concrete with one and two layers of CFRP wraps.

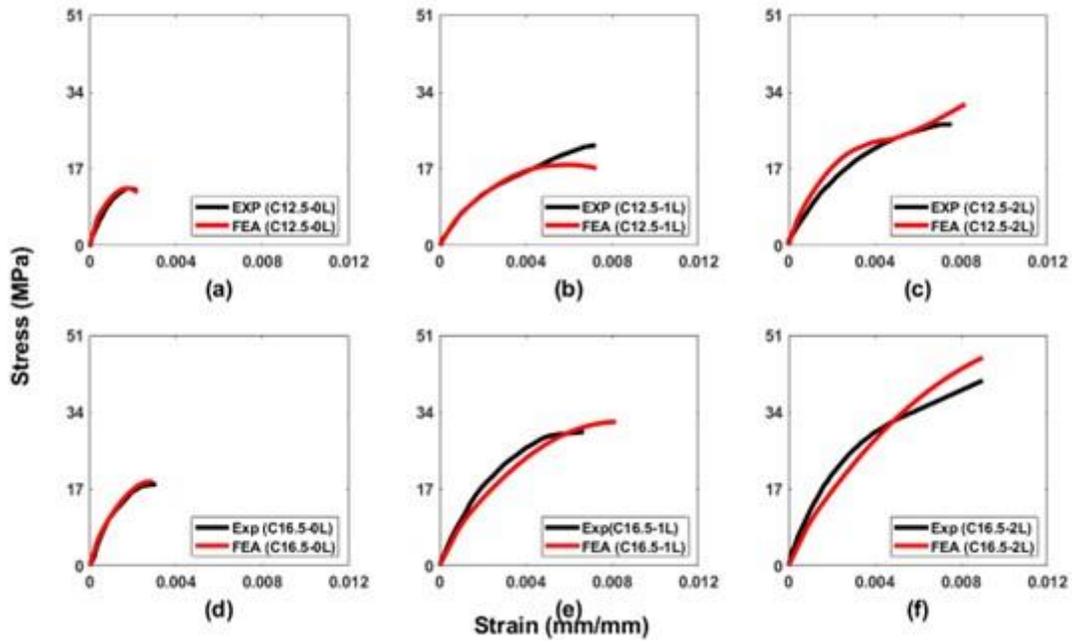

**Figure 15.** Stress-Strain plots for unconfined and confined samples.

## 7 Parametric Study

The influence of each of the individual experimental parameters on the uniaxial compressive strength was determined by the parametric study. In a parametric study, various unconfined concrete strengths ($f'_{co}$) were considered and their effect on the confined strength of concrete ($f'_{cc}$) was analyzed using a proposed model. The study found that an increase in $f'_{co}$ from 5 MPa to 50 MPa, and an increase in the diameter of the cylinder from 100 mm to 550 mm at constant fiber elastic modulus ($E_f$), led to a 193.30% increase in confined strength. When the elastic modulus of FRP increased from 110 GPa to 245 GPa with the same increment of $f'_{co}$, the confined strength increased by 484.91%. Additionally, increasing the thickness of the FRP wrap from 0.15 mm to 1.05 mm while also increasing $f'_{co}$ from 5 MPa to 50 MPa resulted in a significant 761.75% increment in $f'_{cc}$. These findings suggest that the increase in $f'_{co}$ has a more dominant effect when the thickness of the FRP wrap is increased. With an increase in the elastic modulus of FRP from 110 GPa to 245 GPa and an increase of FRP thickness from 0.15mm to 1.05 mm, $f'_{cc}$ increased by 55.75%. It has been observed that there was a negligible effect of an increase in the thickness of FRP layers in large-diameter concrete samples. The thickness of FRP layers significantly influenced the samples with a smaller diameter. With the increase in diameter of the concrete cylinders, the effect of elastic modulus on the confined compressive strength is reduced.

### 7.1 Concrete Compressive Strength and Cylinder Diameter

The most important geometric parameter for the comparison is the diameter of the cylinder and the important mechanical property is the compressive strength of the concrete as illustrated in Figure 16 and 17. These two parameters have a higher influence over the confined compressive strength of the CFRP confined cylinders. The comparative study of the diameter and unconfined compressive strength shows that PSO and GWO have performed better for prediction models and the empirical models of the Lam and Teng and Miyauchi have also provided good results. The Lam and Teng model and Miyauchi model have provided results closer to the unity line than the prediction models. BA has resulted in greater deviations from the unity line.

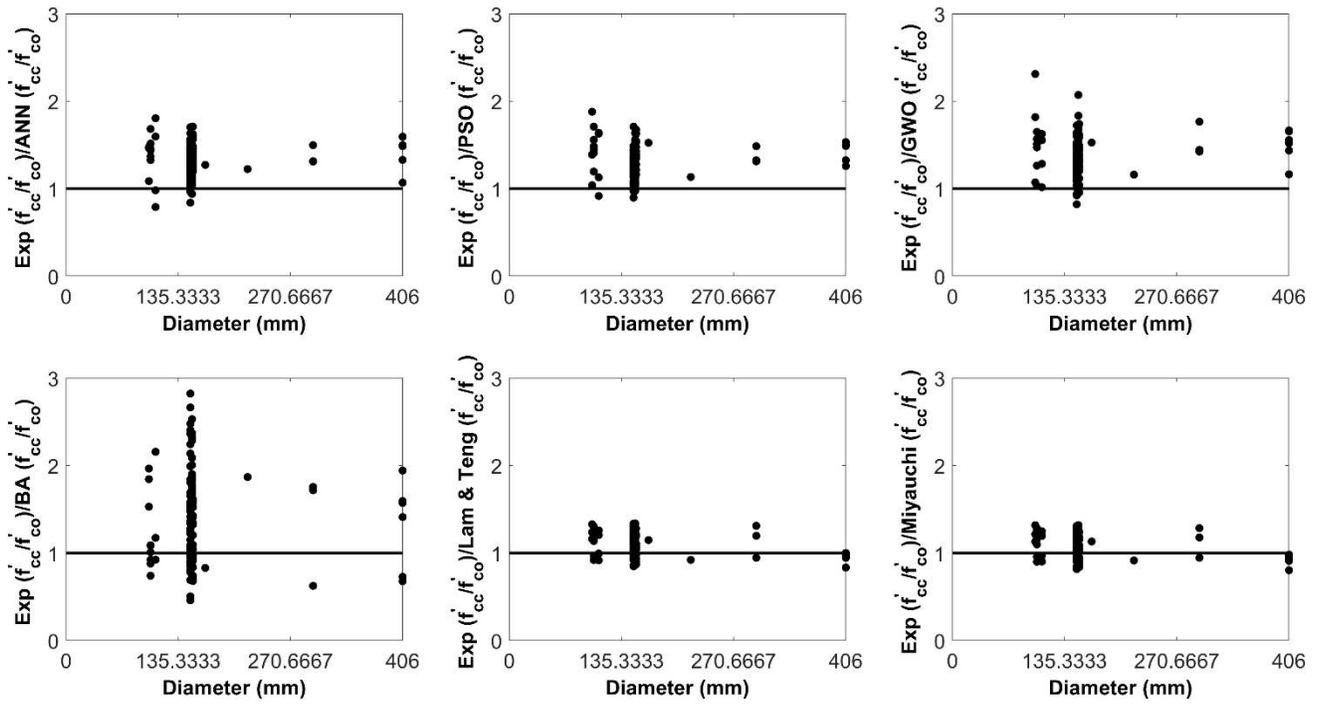

**Figure 16.** Ratio-diameter plots.

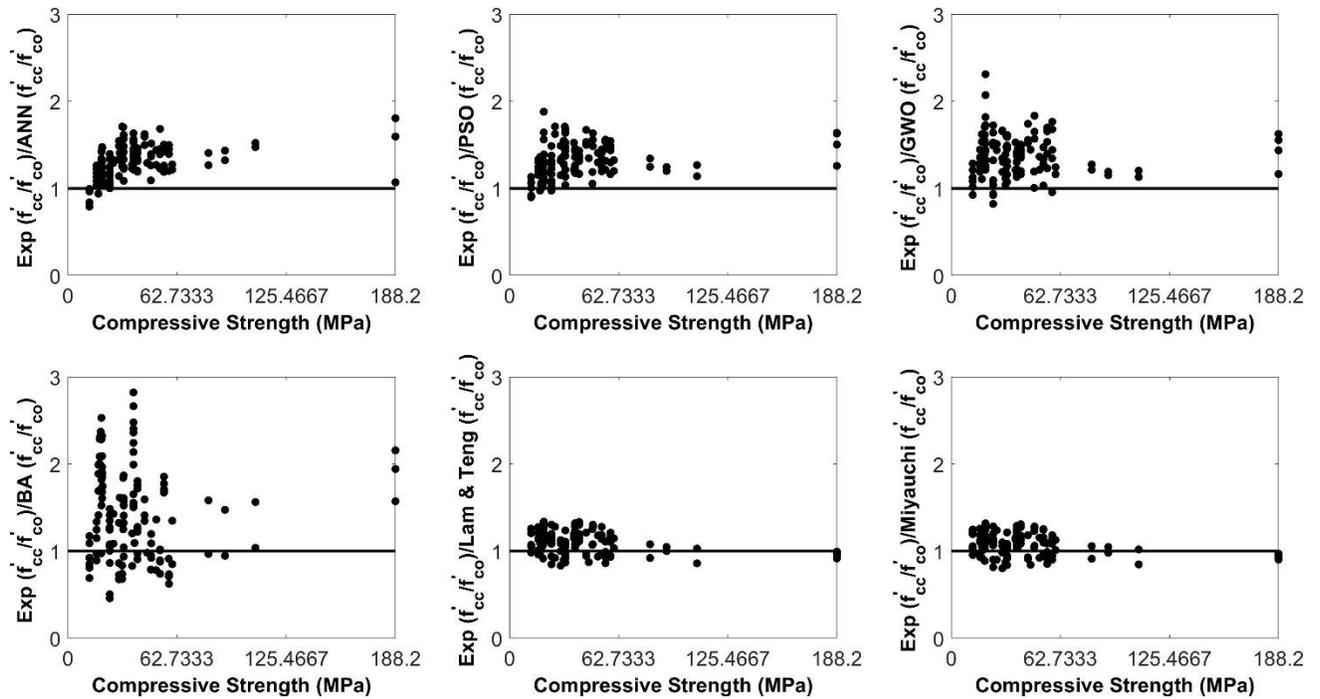

**Figure 17.** Ratio-compressive strength plots.

## 7.2 Elastic Modulus and FRP Thickness

The elastic modulus of the FRP and its thickness is the most important material property related to the FRP as illustrated in Figure 18 and 19. The prediction results of the PSO are closer to the unity line for the Elastic modulus and the thickness of the FRP. The scatter of the points for the BA displayed extensive deviation from the line of unity. While the empirical models of the Lam and Teng model and Miyauchi model indicated a lesser deviation of the results of the ratio of strength from the unity line. The prediction models of the PSO, GWO, and ANN show little bias towards underestimation of the ratio of the $f'_{cc}/f'_{co}$, as depicted in Figure 18 and 19. The results of their ratios are scattered more above the unity line while the results of the Lam and Teng model and Miyauchi model are more uniformly distributed around the unity line that demonstrates the non-biases of the empirical model towards either overestimation or underestimation for the ratio of $f'_{cc}/f'_{co}$.

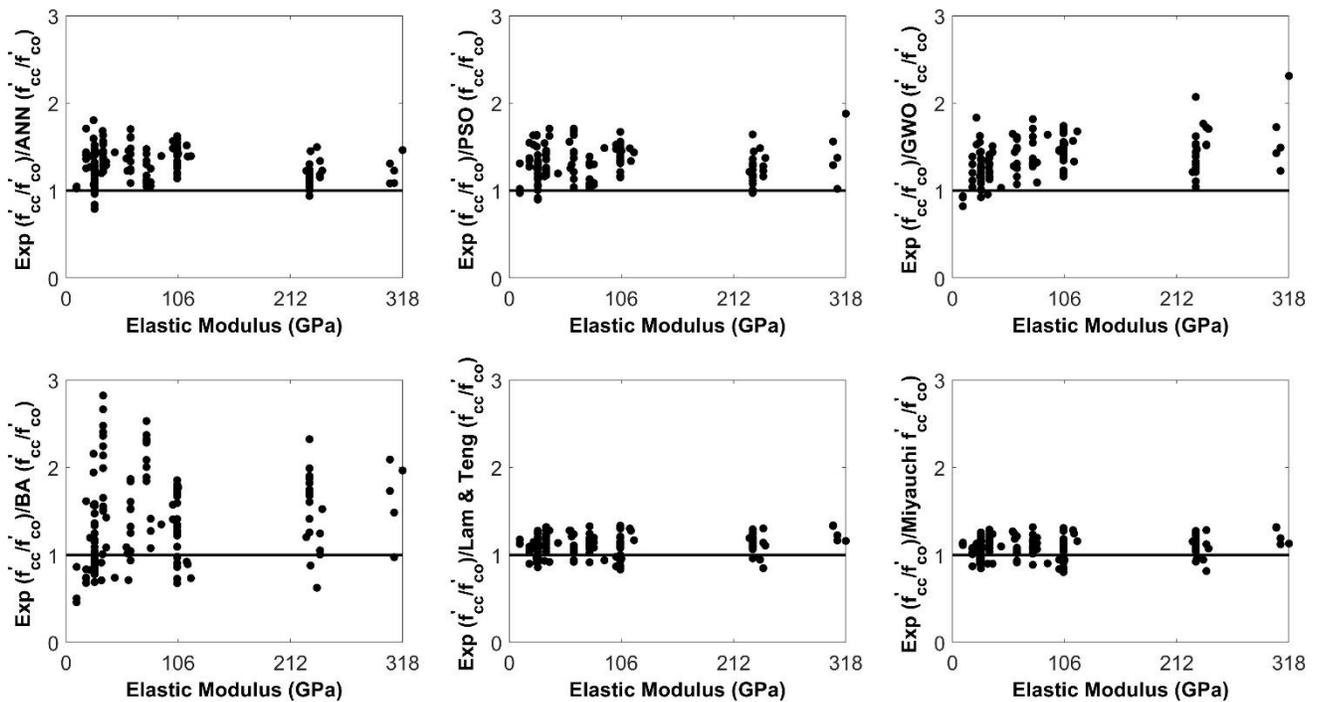

**Figure 18.** Ratio-elastic modulus plots.

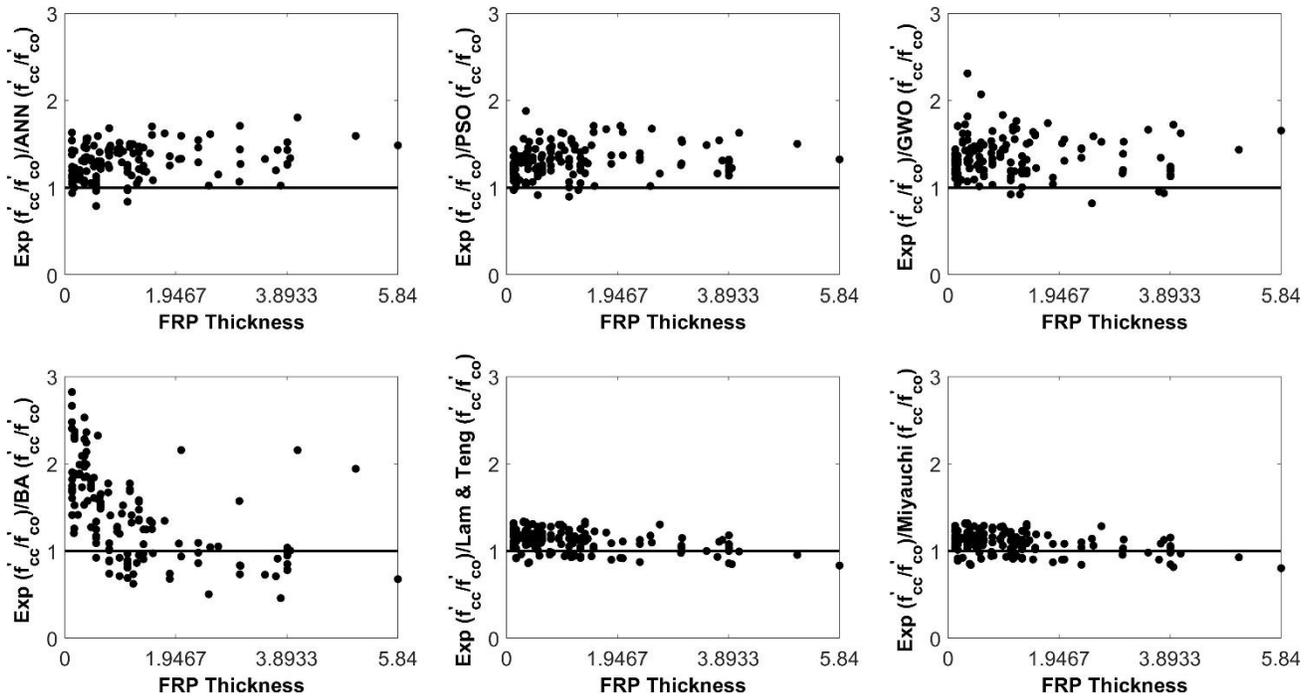

**Figure 19.** Ratio-FRP thickness plots.

## 8 Conclusions

This study evaluated the improvement in the strength of concrete cylinders confined with CFRP and the effectiveness of the optimization ability of metaheuristics models of PSO, GWO, and BA combined with ANN to predict the axial compressive strength of CFRP-confined concrete cylinders. The main conclusions from the research are given below.

The hybrid ANN model of PSO performed the most accurate predictions for the strength of CFRP confined cylinders with the highest $R^2$ of 99.13% and the least *MSE* and *MAE* of 0.14% and 1.28%. The GWO model provided the second most accurate results with an accuracy of 98.17%. The improved accuracy of prediction models suggests that the axial compressive strength of CFRP-confined concrete can be determined accurately. These prediction models can be used as a non-destructive approach to determining the strength of CFRP-confined concrete.

The empirical models of the Lam and Teng model and Miyauchi model provided good results, but they have a bias towards overestimation of the values. Therefore, the prediction plots showed the maximum deviation of the best-fit line of the calculated results from the ideal best-fit line. BA performed the least accurate predictions but still, its best-fit line is closer to the ideal line than that of the empirical models.

The experiments performed by the authors in the previous research demonstrated that confining concrete cylinders with CFRP increased the axial compressive strength and ductility of the concrete cylinders. The strength of the confined concrete enhanced with the increase in the unconfined strength, the elastic modulus of CFRP, and the thickness of CFRP layers.

The performance of three hybrid models of PSO, GWO, and BA clarified that not all metaheuristics-based hybrid ANN models could outperform ANN. PSO performance based on the predictions and error metrics revealed that it performed better than ANN but under the same training and testing parameters, BA performed less accurately as compared to ANN. Furthermore, the performance of the metaheuristics-based ANN models varied with the objective functions. The parameters of the hybrid models were optimized manually through careful observation of training and testing results of the models by varying the parameters of the models. The performance could be improved by using the automatic process of parameter optimization (Neumüller et al. 2012).

**Funding:** This research received no external funding.

**Conflicts of Interest:** The authors declare no conflict of interest.